\documentclass{midl} % Include author names

\usepackage{booktabs}
\usepackage{multirow}
\usepackage{adjustbox}
\usepackage{array}

\makeatletter
\renewcommand*\@jmlrpages{}
\makeatother

\title[CardAIc-Agents for Cardiac Care Support]{CardAIc-Agents: A Multimodal Framework with Hierarchical Adaptation for Cardiac Care Support}

\midlauthor{\Name{Yuting Zhang\nametag{$^{1}$}} \Email{ytz300@student.bham.ac.uk}\\
\addr $^{1}$ School of Computer Science, University of Birmingham, UK \\
\Name{Karina V. Bunting\nametag{$^{2}$}} \Email{k.v.bunting@bham.ac.uk}\\
\Name{Asgher Champsi\nametag{$^{2}$}} \Email{a.champsi@bham.ac.uk}\\
\addr $^{2}$ Department of Cardiovascular Sciences, University of Birmingham, UK \\
\Name{Xiaoxia Wang\nametag{$^{2,3}$}} \Email{xiaoxia.wang@uhb.nhs.uk}\\
\addr $^{3}$ NIHR Birmingham Biomedical Research Centre and West Midlands NHS Secure Data Environment, University Hospitals Birmingham NHS Foundation Trust, UK \\
\Name{Wenqi Lu\nametag{$^{4}$}} \Email{W.Lu@mmu.ac.uk}\\
\addr $^{4}$ Department of Computing and Mathematics, Manchester Metropolitan University, UK \\
\Name{Alexander Thorley\nametag{$^{1}$}} \Email{ajt973@student.bham.ac.uk}\\
\Name{Sandeep S Hothi\nametag{$^{5}$}} \Email{s.hothi@nhs.net}\\
\addr $^{5}$ Department of Cardiology, Heart and Lung Centre, Royal Wolverhampton NHS Trust, UK \\
\Name{Zhaowen Qiu\nametag{$^{6}$}} \Email{249600398@qq.com}\\
\addr $^{6}$ College of Computer and Control Engineering, Northeast Forestry University, China \\
\Name{Baturalp Buyukates\midlotherjointauthor\nametag{$^{1}$}} \Email{b.buyukates@bham.ac.uk}\\
\Name{Dipak Kotecha\midlotherjointauthor\nametag{$^{2,3,7}$}} \Email{d.kotecha@bham.ac.uk}\\
\addr $^{7}$ Julius Center, University Medical Center Utrecht, the Netherlands\\
\Name{Jinming Duan\midlotherjointauthor\nametag{$^{1,8}$}} \Email{j.duan@bham.ac.uk}\\
\addr $^{8}$ Division of Informatics, Imaging and Data Sciences, University of Manchester, UK
}

\begin{document}
\maketitle
\begin{abstract}
Cardiovascular diseases (CVDs) remain the foremost cause of mortality worldwide, a burden worsened by a severe deficit of healthcare workers. Artificial intelligence (AI) agents have shown potential to alleviate this gap through automated detection and proactive screening, yet their clinical application remains limited by: 1) rigid sequential workflows, whereas clinical care often requires adaptive reasoning that select specific tests and, based on their results, guides personalised next steps; 2) reliance solely on intrinsic model capabilities to perform role assignment without domain-specific tool support; 3) general and static knowledge bases without continuous learning capability; and 4) fixed unimodal or bimodal inputs and lack of on-demand visual outputs when clinicians require visual clarification. In response, a multimodal framework, CardAIc-Agents, was proposed to augment models with external tools and adaptively support diverse cardiac tasks. First, a CardiacRAG agent generated task-aware plans from updatable cardiac knowledge, while the Chief agent integrated tools to autonomously execute these plans and deliver decisions. Second, to enable adaptive and case-specific customization, a stepwise update strategy was developed to dynamically refine plans based on preceding execution results, once the task was assessed as complex. Third, a multidisciplinary discussion team was proposed which was automatically invoked to interpret challenging cases, thereby supporting further adaptation. In addition, visual review panels were provided to assist validation when clinicians raised concerns. Experiments across three datasets showed the efficiency of CardAIc-Agents compared to mainstream Vision–Language Models (VLMs) and state-of-the-art agentic systems. Code will be publicly available at \url{https://github.com/ytz300/CardAIc-Agents}.
\end{abstract}

% 1.check
% First, a CardiacRAG agent employed the proposed hybrid retrieval mechanism and generated task-aware plans from the updatable cardiac knowledge base.

% a multidisciplinary discussion team was proposed to be automatically invoked for challenging cases, thereby supporting further adaptation.


\section{Introduction}
Cardiovascular diseases (CVDs) are the leading cause of mortality worldwide, accounting for 17.9 million deaths each year \cite{almeida2024cardiovascular}. Notably, up to 80\% of these deaths occur in low- and middle-income countries, where specialised care is limited \cite{bulto2024burden}, which, combined with a global shortage of over 4 million healthcare workers \cite{vedanthan2011urgent}, drives the demand for scalable and accessible cardiovascular care solutions. Recent advancements in large language models (LLMs) have ave led to human-level performance on challenging tasks; for instance, Med-PaLM has outperformed clinicians on the United States Medical Licensing Examination \cite{singhal2025toward}. Despite these achievements, however, clinical practice, particularly for complex chronic conditions (e.g., heart failure (HF)), often relies on multimodal data for diagnosis, prognosis, and treatment \cite{weintraub2019role}. This gap underscores the need for multimodal strategies that extend beyond language-only models to more effectively support clinical practice.

While vision-language models (VLMs) such as LLaVA-Med \cite{li2023llavamed} and MedGe- mma \cite{sellergren2025medgemma} have fueled anticipation for medical multimodal artificial intelligence (AI), several challenges remain. For example, they are restricted to static images, whereas dynamic inputs such as echocardiograms are vital for cardiac function assessment. In addition, such generalist models retain static knowledge, which hinders their ability to assimilate evolving medical evidence. While Retrieve Augmented Generation (RAG) mitigates this challenge to some extent, traditional retrieval methods still present notable limitations. For example, Term Frequency Inverse Document Frequency (TF-IDF) relies on lexical matching but is limited in semantic comprehension, while Dense Passage Retrieval (DPR) encodes queries and documents into embeddings for similarity based retrieval yet often lacks semantic relevance \cite{karpukhin2020dense, mallen2022not}.

Crucially, complex cardiovascular management often requires multi-step reasoning and a coordinated sequence of clinical actions, rather than a single-step response \cite{mcdonagh20212021}. Although prompt engineering techniques such as Chain-of-Thought (CoT) \cite{wei2022chain} partially mitigate this limitation by decomposing problems into substeps, model performance remains constrained by their intrinsic capabilities. The recent introduction of function calling and the Model Context Protocol (MCP) \cite{hou2025model} provides a complementary pathway, enabling models to integrate external tools automatically and access standardized functions. These advances drive the development of AI agents capable of reasoning, planning, memory utilization, and action execution \cite{chang2024agentboard}. However, most existing VLM-based agents in medicine still rely on assigning roles to models with static and generic knowledge, and often lack cross-turn memory, limiting their suitability for real-world cases that require multidisciplinary deliberation.  

Another limitation of these existing VLM-based agents lies in their rigid and sequential workflows \cite{kim2024mdagents}. Although recent advances have enabled ReAct-based frameworks \cite{yao2023react} to perform stepwise reasoning with intermediate outcomes and external tools, these frameworks still lack global planning capability. In contrast, MedAgent-Pro \cite{wang2025medagent} offers disease-level planning, but its plans are generated before receiving patient-specific input, which may be effective for some routine tasks yet risks misalignment with individual clinical contexts. For example, echocardiogram view identification typically follows a fixed workflow (e.g., commercial software, manual view selection), whereas complex HF diagnosis involves diverse patient presentations that require tailored test orders (e.g., ECG, echocardiography) and subsequent personalised management based on results. These observations highlight the need for flexible frameworks capable of both task-level and case-level adaptation across diverse clinical contexts. In addition, current VLM agents lack intermediate visual outputs, such as  the left ventricular contour delineations, which are critical for clinical verification in complex or uncertain cases.

Motivated by the above, in this study, an adaptive framework, CardAIc-Agents (comprising CardiacRAG and a Chief agent), was introduced to augment models with external tools, enabling autonomous execution of cardiac tasks (e.g., diagnosis, echocardiogram view extraction, segmentation, detection of P, QRS, and T waves) across diverse modalities (e.g., textual, signal, image, and video). Specifically, the CardiacRAG agent was developed to formulate general plans based on the latest domain knowledge and the proposed hybrid retrieval technique, whereas the Chief agent enhanced its own capabilities through the integration and orchestration of external tools for plan execution and definitive decision-making. To support adaptive planning across tasks and patient-specific cases, the system initially assessed task complexity, executed the plan, and dynamically refined it as new evidence emerged. For more challenging cases, a multidisciplinary discussion team (MDT), augmented with external tools and cross-turn memory, was proposed to support further interpretation. Finally, when clinicians raised concerns, visual review panels were provided for validation. In summary, the key contributions can be articulated as follows:

\begin{itemize}    
    \item A domain-specific framework, CardAIc-Agents, was developed to enhance the capabilities of large models through specialized tool integration, enabling autonomous execution of diverse cardiac tasks on multimodal data.
    \item Adaptive strategies were proposed to stratify task complexity, refine plans iteratively as new evidence emerged, initiate team discussions, and provide visual validation, enabling hierarchical adaptation tailored to specific tasks and individual patients.
    \item A CardiacRAG agent was introduced to derive plans based on an updated cardiac knowledge base, while employing a hybrid retrieval mechanism to optimize semantic comprehension and relevance. 
    \item A multidisciplinary discussion team was designed to integrate external tools that extend the static and general capabilities of foundation models and to incorporate cross-turn memory that preserves context across reasoning steps.
    
\end{itemize}  

\section{Method}
\begin{figure}[t!]
%\floatconts
\centering
\includegraphics[width=0.85\linewidth]{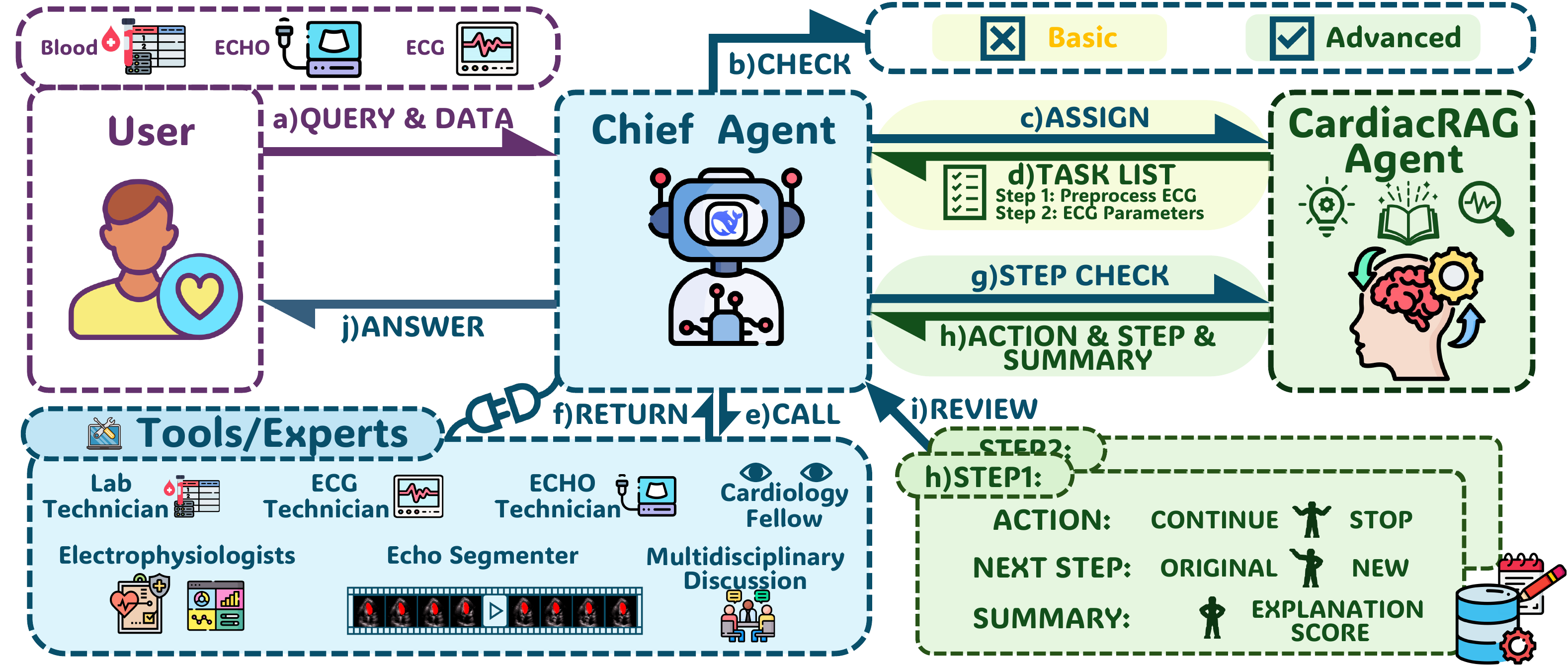}
\caption{Overview of the CardAIc-Agents framework.}
\label{fig:workflow}
\vspace{-5mm}
\end{figure}

CardAIc-Agents consist of two components: the CardiacRAG and the Chief agent (Figure~\ref{fig:workflow}). The former, based on a dedicated cardiac knowledge base, generates and updates plans as new evidence emerges. The Chief agent serves as the primary decision-maker, responsible for complexity assessment, task assignment, plan execution, and tool invocation.

%CardAIc-Agents consist of two components: the CardiacRAG and the Chief agent (Figure \ref{fig:workflow}). 
As shown in Figure~\ref{fig:workflow}, upon receiving the query with associated multimodal data such as ECGs (a), the Chief agent performs a complexity assessment (b) and assigns the task to the CardiacRAG agent (c), which retrieves domain-specific evidence from a curated cardiac knowledge base and constructs a general plan for the case (d). For low-complexity cases, the Chief agent executes this plan by invoking the required analytical tools (e), integrates their results (f), and generates the final clinical response (j). For high-complexity cases, the plan is adaptively revised (g, h) as new evidence (i) is incorporated, enabling iterative refinement of subsequent reasoning and tool selection before the final decision is produced (j). For more challenging cases, the Chief agent may autonomously initiate a multidisciplinary discussion team (MDT) to provide enhanced interpretation. When clinicians raise concerns, visual review panels may be required as an available option for human validation.

\subsection{CardiacRAG Agent}
\begin{figure}[t!]
%\floatconts
\centering
\includegraphics[width=0.75\linewidth]{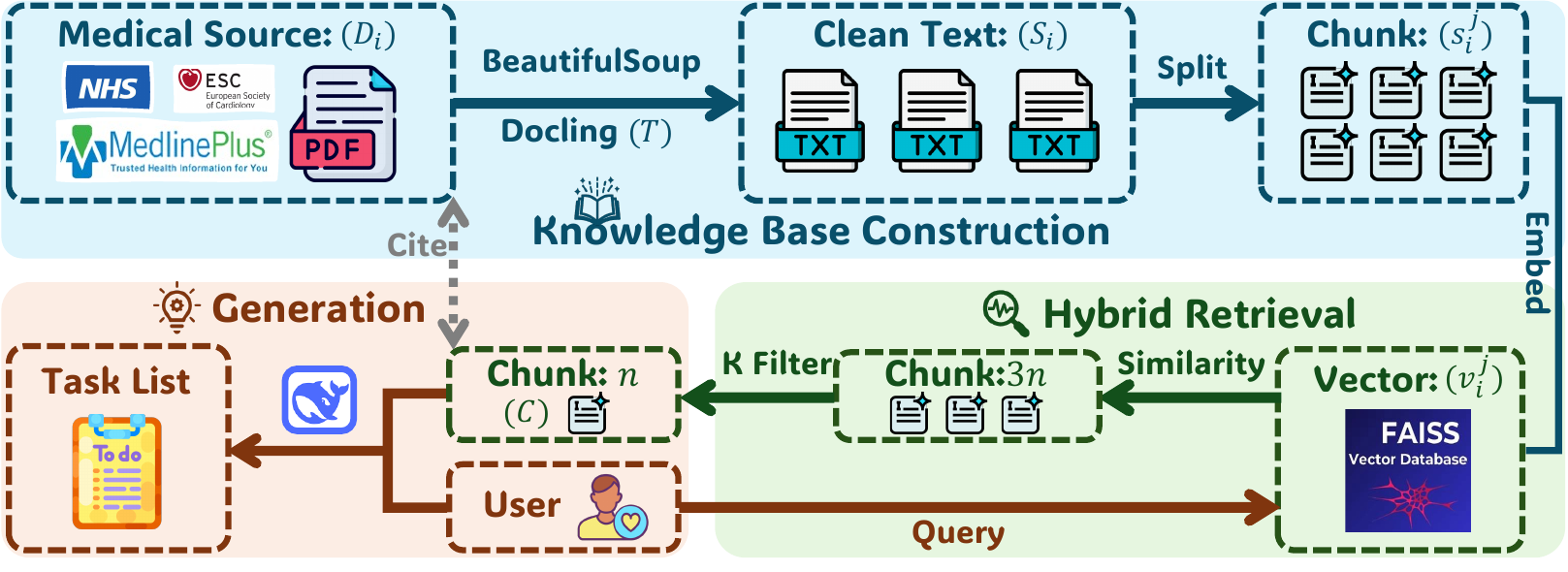}
\caption{Illustration of the CardiacRAG agent. $D_i$ denotes the \(i\)-th source, \(S_i\) is the cleaned text, \(s_i^j\) is the \(j\)-th chunk from source \(i\), \(v_i^j\) is its corresponding vector, \(T\) represents the transformation method, \(K\) denotes keyword-based filtering, \(n\) is the number of chunks retrieved, \(c\) is the final retrieved content, and \textit{Cite} indicates optional return of original chunks for transparency and reference.}
\label{fig:kb}
\vspace{-5mm}
\end{figure}

The CardiacRAG agent was developed as current LLMs and VLMs encode static knowledge and their general-purpose design may lack cardiovascular domain specificity. This agent emulates the clinician reasoning process through information retrieval from authoritative medical sources and it was structured into three key stages (see Figure \ref{fig:kb}).

\smallskip
\noindent\textit{Knowledge base construction.} 
To reduce the complexity of information retrieval and improve accuracy, the knowledge base construction process focused exclusively on cardiac content. This domain-specific approach selectively aggregated data \(\{D_i\}_{i=1}^M=\{ D_1, D_2, \ldots, D_M \}\) from authoritative medical sources, including major US academic medical centers (e.g., Mayo Clinic \citeyear{mfmmer2025}), UK National Health Service (\citeyear{nhs2025}), health information platforms (e.g., MedlinePlus \citeyear{miller2000medlineplus}), and recently published official guidelines (see the appendices for details).

Then, the raw documents \(\{D_i\}_{i=1}^M\) were preprocessed through a transformation function \(T\) that extracts and normalizes textual content, producing clean text:
\begin{equation}
\{S_i\}_{i=1}^M = T\big(\{D_i\}_{i=1}^M\big), \quad i = 1,2,\ldots,M.
\end{equation}
where \(T\) refers to BeautifulSoup  \cite{abodayeh2023web} for HTML files and Docling  \cite{livathinos2025docling} for PDFs, and \(M\) refers to the total number of collected documents. 

Finally, cleaned texts $S_i$ were split into chunks \( s_i^j \) to preserve contextual continuity:
\begin{equation}
s_i^j = \text{chunk}_j(S_i; d_s, d_o), \quad j = 1, 2, \ldots, L_i,
\end{equation}
where \(L_i\) is the chunk count of document \(i\); \(j\) is the chunk index; and $d_s$ and $d_o$ denote the chunk and overlap size, respectively.

% where \( L_i \) denotes the total number of chunks generated from document \( i \); % \( j \) indexes the chunk position within document \( i \); \( d_s \) and \( d_o \) represent the chunk and overlap size, respectively.
% Both were hyperparameters in this work.

\smallskip
\noindent\textit{Hybrid retrieval.} 
To reduce irrelevant results from current vector similarity retrieval techniques, a hybrid retrieval method, combined with TF-IDF variants, was applied to further filter results using domain-specific keywords and ensure clinical specificity (see Figure \ref{fig:kb}).

{\scriptsize{$\bullet$}} Vector similarity retrieval. To preserve semantic relevance, both chunks \( s_i^j \) and the query were embedded via Bio\_ClinicalBERT \cite{alsentzer-etal-2019-publicly} as \(v_i^j=\phi(s_i^j)\) and \( q=\phi(\text{query}) \). The set of document vectors $\mathcal{V}=\bigcup_{i=1}^{M} \{ v_i^j \}_{j=1}^{L_i}$ was ranked by cosine similarity:
\begin{equation}
\quad
\mathrm{sim}(q, v_i^j) = \frac{q \cdot v_i^j}{\|q\| \|v_i^j\|},
\end{equation}
and the top \(3n\) vectors (\(\mathcal{V}_{(1)},\ldots,\mathcal{V}_{(3n)}\)) were returned, where \( n \) is the final number of results. Document vectors were indexed and stored in the FAISS vector database for efficient retrieval and reused without recalculation in subsequent queries.

{\scriptsize{$\bullet$}} Keyword-based filtering. To improve clinical relevance, top \(3n\) documents were
further filtered based on domain-specific weights:
\begin{equation}
MW(k) = 
\begin{cases} 
\omega_{\text{medical}}[k], & k \in \mathcal{B}_{\text{medical}} \\
1, & \text{otherwise}
\end{cases},
\end{equation}
where \(k\) is the keyword from the query \(Q\), \(\mathcal{B}_{\text{medical}}\) is the clinical vocabulary, and \(\omega_{\text{medical}}\) is the term importance defined based on the clinical context. To exploit structural cues, a position-based bonus \cite{hofstatter2021mitigating} was introduced: 

%self-defined or clinical domain knowledge-based
\begin{equation}
PB(k,s_i^j) = 
\begin{cases} 
1.2, & \text{if } \text{pos}(k,s_i^j) < 0.3 \times d_s \\
1, & \text{otherwise}
\end{cases},
\end{equation}
where \(\text{pos}(k,s_i^j)\) is the first index of \(k\) in chunk \(s_i^j\). The final retrieval score then became:
\begin{equation}
\text{Score}_{(s_i^j,Q)} = \frac{1}{|Q|} \sum_{k \in Q} TF(k,s_i^j) \cdot MW(k) \cdot PB(k,s_i^j),
\end{equation}
where chunks with scores above threshold \(\theta\) were retained, and term frequency defined as:
\begin{equation}
TF(k,s_i^j) = \frac{\text{count}(k,s_i^j)}{|\text{words}(s_i^j)|}.
\end{equation}

\smallskip
\noindent\textit{Guideline generation.} 
Based on the retrieved chunks \( C=\{c_1, c_2, \dots, c_n\} \) and the query $Q$, the general plan was generated by the CardiacRAG agent, which employed DeepSeek-R1-Distill-Qwen-32B \cite{deepseekai2025deepseekr1incentivizingreasoningcapability} as its core model. This general plan could be continuously updated as new results become available during subsequent steps, thereby reflecting clinical practices. 

% DeepSeek-R1-Distill-Qwen-32B \cite{deepseekai2025deepseekr1incentivizingreasoningcapability} was employed within this CardiacRAG agent. 

{\scriptsize{$\bullet$}} General plan. Given the query \(Q\), the model did not reconstruct the knowledge base; instead, it retrieved relevant evidence from the prebuilt cardiac knowledge base (FAISS vector database described earlier) using the hybrid retrieval mechanism. The retrieved chunks \(C\), together with the query, provided the contextual input from which the model generated an initial stepwise plan (see the appendices for the prompt):
\begin{equation}
P = \texttt{DeepSeek-R1}(\texttt{PlanPrompt}(Q, C)),
\end{equation}
where the plan \(P=(p_1,\ldots,p_s) \) contains \( s \) steps, which may vary by case. 

% where \texttt{PlanPrompt} formats the query and chunks for procedural guidance generation (see the appendices \ref{app:prompt} for the prompt). The plan \(P=(p_1,\ldots,p_s) \) contains \( s \) steps, which may vary by case. 

{\scriptsize{$\bullet$}} Stepwise update. At each step, the CardiacRAG agent evaluated the execution state and determined whether the next action should be revised based on intermediate evidence. Formally, at step $i$, the model received the execution history $\log_i$ and the proposed next step $p_{i+1}$, and produced an updated decision:
\begin{equation}
(S_i, A_i, p_{i+1}')
= \texttt{DeepSeek-R1}\big(\texttt{UpdatePrompt}(\log_i, p_{i+1})\big),
\qquad i = 1,\ldots,s-1.
\end{equation}
Here, $S_i$ summarizes the evidence, $A_i \in \{\text{stop},\text{continue}\}$ specifies whether execution terminates, and $p_{i+1}'$ is the updated next step; if no revision is required, $p_{i+1}' = p_{i+1}$. The procedure halts at the first index $k$ such that $A_k = \text{stop}$, or proceeds through all $s$ steps otherwise. A detailed case study is provided in the appendices.

\subsection{Chief Agent}
The Chief agent leverages the advanced reasoning capabilities of DeepSeek-R1 to coordinate specialized tools and apply adaptive strategies that adjust behaviour at both the task and patient levels, reflecting clinical workflows and supporting real-world application.

\smallskip
\noindent\textit{Adaptive strategies.}
Given the query $Q$, the Chief agent first assessed the task complexity:
\begin{equation}
\ell = \texttt{DeepSeek\_R1}(\texttt{Prompt}(Q)),
\qquad 
\ell \in \{\text{basic}, \text{advanced}\}.
\end{equation}
% The used prompt is given in the appendices \ref{app:prompt}.

Based on the predicted complexity level \( \ell \), the agent executed either the general plan or a stepwise refinement procedure. Execution in both modes follows:
\begin{equation}
p_{i+1}^{*} =
\begin{cases}
p_{i+1}, & \ell=\text{basic},\\[3pt]
p_{i+1}', & \ell=\text{advanced},
\end{cases}
\qquad i = 1,\ldots,k.
\end{equation}

At each step, the Chief automatically executed the actual action 
\(p_{i+1}^{*}\) by invoking tool \(T_{i+1} \in \mathcal{T}\) (see the appendices for the used tools), yielding the output
\(t_{i+1} = T_{i+1}(p_{i+1}^{*})\). For complex cases, the Chief may additionally invoke the MDT tool to simulate clinical case conferences. The evidence was then appended to the log via \(\log_{i+1}=\texttt{Append}(\log_i, t_{i+1})\). After all steps, the Chief synthesized the log to generate the final summary and answer:
\begin{equation}
(\text{Summary}, \text{Answer}) = \texttt{DeepSeek-R1}(\log).
\end{equation}

Further, the overall system could provide visual validation when required for disputed or ambiguous cases (see Figure \ref{fig:panel}), allowing clinicians to perform manual verification:
\begin{equation}
V = \texttt{CardAIc-Agents}(Q),
\end{equation}
which encapsulates the above workflow before producing the final visual output \(V\).

\begin{figure}
    %\floatconts
    \centering
    \includegraphics[width=1\linewidth]{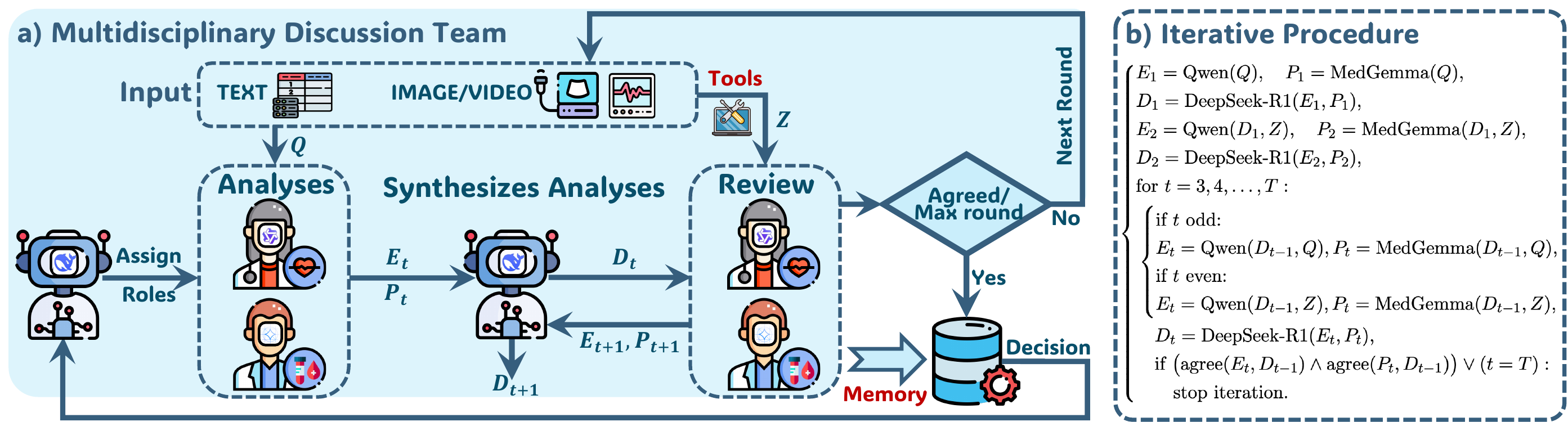}
    \caption{(a) Illustration of the Multidisciplinary Discussion Team. (b) Iterative Procedure. \( Q \) denotes the original input, \( Z \) the intermediate tool outputs, \( E_{*}, P_{*}, D_{*} \) the model responses, with \( T \) as the total steps and \( t \) the step index.}  
    \label{fig:team}
    \vspace{-5mm}
\end{figure}

\smallskip
\noindent\textit{Multidisciplinary discussion team (MDT).} As shown in Figure \ref{fig:team}, this team reviewed inputs and intermediate outputs from tools to support comprehensive decisions. First, the Chief designated two relevant domain-expert roles based on the inputs. Each expert independently analyzed the inputs, and their respective results were synthesized by the Chief. The two experts then reviewed this synthesis together with outputs produced by tools, enabling the integration of information sources that extended beyond the large model review paradigm typically adopted in existing medical agents. The Chief subsequently re-synthesized the updated information, completing one discussion round. All intermediate results generated during each round were stored as memory for downstream use.

If consensus was reached, or the maximum number of predefined discussion rounds was exhausted, the Chief issued final decision (see the appendices). Otherwise, the synthesized output from the current round was combined with the original inputs and passed to the next round, ensuring that subsequent iterations built upon accumulated evidence rather than relying solely on the initial inputs or on the stochastic behavior of responses from LLMs or VLMs. In this study, the two experts were implemented using MedGemma \cite{sellergren2025medgemma}, specialized for medical image analysis, and Qwen2.5-VL \cite{qwen2.5-VL}, specialized for video processing as noted in Figure~\ref{fig:team}(b).

% laboratory test results (Labs), 12-lead electrocardiogram (ECGs), and echocardiograms (ECHOs)
\section{Experiments}
\subsection{Experimental Settings} 
\noindent\textit{Datasets.} CardAIc-Agents was evaluated on three datasets: (i) MIMIC-IV \cite{johnson2023mimic}, for HF diagnosis, which includes data from 1,524 patients with three modalities (laboratory test results, 12-lead ECGs, and echocardiograms (ECHOs));  (ii) PTB-XL \cite{wagner2020ptb}, for Myocardial infarction (MI) diagnosis \cite{strodthoff2023ptb}, which includes data from 10,147 patients with structured patient information, ECG-derived variables, and 12-lead ECGs; (iii) The PTB Diagnostic ECG Database contains patient information and 12-lead ECGs from 268 cases for HF prediction (see the appendices for details).

\smallskip
\noindent\textit{Metrics and baseline.}
The diagnostic performance was evaluated using the area under the receiver operating characteristic curve (AUC) and accuracy \cite{yu2021evaluation}, with 95\% confidence intervals. For the visual outputs, two cardiologists independently assessed and scored the results. The proposed agent was compared with medical VLMs, including LLaVA-Med \cite{li2023llavamed} and MedGemma \cite{sellergren2025medgemma}, with MedGemma combined with CoT \cite{wei2022chain} and ReAct \cite{yao2023react} to evaluate step-by-step reasoning and tool-augmented strategies. Comparisons were also made with medical agent frameworks such as MedAgents \cite{tang2024medagents}, ReConcile \cite{chen-etal-2024-reconcile}, and MDAgents \cite{kim2024mdagents}. Further implementation details and additional experiments are given in the appendices, respectively.

\subsection{Results and Comparative Analysis}
\begin{table*}[t]
\floatconts
  {tab:cp}% label
  {\caption{Performance Comparison Across Methods and Datasets}}% caption
  {%
    \centering
    \vspace{-8pt}
    %\small
    \setlength{\tabcolsep}{2.5pt}
    \begin{adjustbox}{width=\textwidth}
    \begin{tabular}{cccccccc}
    \toprule
    \multirow{2}{*}{\textbf{Category}} & 
    \multirow{2}{*}{\textbf{Method}} & 
    \multicolumn{2}{c}{\textbf{MIMIC-IV}} & 
    \multicolumn{2}{c}{\textbf{PTB-XL}} & 
    \multicolumn{2}{c}{\textbf{PTB Diagnostic}} \\
    \cmidrule(lr){3-4} \cmidrule(lr){5-6} \cmidrule(lr){7-8}
    & & \textbf{ACC} & \textbf{AUC} & \textbf{ACC} & \textbf{AUC} & \textbf{ACC} & \textbf{AUC} \\
    \midrule

    \multirow{4}{*}{\textbf{\begin{tabular}[c]{@{}c@{}}VLMs \\ \&\\Variants\end{tabular}}} 
    & LLaVA-Med(\citeauthor{li2023llavamed}) & 0.35 (0.30, 0.41) & 0.34 (0.28, 0.40) & 0.39 (0.37, 0.42) & 0.51 (0.47, 0.55) & 0.12 (0.08, 0.16) & 0.44 (0.34, 0.59) \\
    & MedGemma(\citeauthor{sellergren2025medgemma}) & 0.76 (0.71, 0.80) & 0.82 (0.77, 0.86) & 0.56 (0.53, 0.59) & 0.55 (0.52, 0.59) & 0.76 (0.71, 0.81) & 0.88 (0.83, 0.93) \\
    & MedGemma(CoT,\citeauthor{wei2022chain}) & 0.65 (0.60, 0.70) & 0.81 (0.76, 0.86) & 0.53 (0.50, 0.56) & 0.54 (0.50, 0.58) & 0.58 (0.52, 0.64) & 0.75 (0.64, 0.88) \\
    & MedGemma(ReAct,\citeauthor{yao2023react}) & 0.67 (0.62, 0.72) & 0.71 (0.66, 0.76) & 0.83 (0.81, 0.85) & 0.83 (0.80, 0.85) & 0.69 (0.64, 0.75) & 0.72 (0.41, 0.99) \\
    \midrule

    \multirow{3}{*}{\textbf{\begin{tabular}[c]{@{}c@{}}Medical\\Agents\end{tabular}}} 
    & MedAgents(\citeauthor{tang2024medagents}) & 0.74 (0.70, 0.79) & 0.82 (0.78, 0.87) & 0.65 (0.62, 0.68) & 0.62 (0.58, 0.66) & 0.75 (0.70, 0.79) & 0.84 (0.74, 0.92) \\
    & ReConcile(\citeauthor{chen-etal-2024-reconcile}) & 0.49 (0.44, 0.55) & 0.75 (0.69, 0.80) & 0.43 (0.40, 0.46) & 0.57 (0.53, 0.61) & 0.55 (0.49, 0.61) & 0.76 (0.58, 0.93) \\
    & MDAgents(\citeauthor{kim2024mdagents})& 0.52 (0.47, 0.58) & 0.61 (0.55, 0.68) & 0.56 (0.52, 0.58) & 0.60 (0.56, 0.64) & 0.74 (0.68, 0.79) & 0.74 (0.53, 0.96) \\
    \midrule

    \textbf{Proposed} & CardAIc-Agents & \textbf{0.87(0.82,0.90)} & \textbf{0.89(0.85,0.93)} & \textbf{0.96(0.95,0.97)} & \textbf{0.96(0.94,0.98)} & \textbf{0.77(0.72,0.82)} & \textbf{0.88(0.65,1.00)} \\
    \bottomrule
    \multicolumn{8}{l}{\normalsize{Note: Boldface values indicate best performance within each dataset and metric; Values in parentheses represent 95\% confidence intervals;}} \\%[-4pt]
    \multicolumn{8}{l}{\normalsize{\phantom{Note: }ACC = accuracy; AUC = Area Under the Curve; CoT = Chain of Thought; ReAct = Reasoning and Acting.}} 
    \end{tabular}
    \end{adjustbox} }
\end{table*}

\noindent\textit{Comparison with VLMs and variants.} As shown in Table \ref{tab:cp}, CardAIc-Agents outperformed all baseline VLMs across the three cardiac datasets. The largest gap was observed on the MIMIC-IV, where CardAIc-Agents achieved an accuracy of 0.87 compared to only 0.35 by LLaVA-Med ($p<0.05$). This was partly due to the limited token input length, which constrained the performance of this medical yet general VLM. Secondly, MedGemma performed the best among the VLMs, while enabling CoT reasoning did not improve performance across all datasets. Finally, its ReAct system, built on LangChain for tool use, improved PTB-XL performance but not MIMIC-IV, and still outperformed by the proposed method.

% Finally, its ReAct system was built on LangChain to invoke tools, which improved performance on PTB-XL but not on MIMIC-IV.
% In summary, the proposed agent outperformed existing general-purpose medical VLMs.

% This disparity likely arised from the higher contextual complexity in MIMIC-IV, which may challenge the VLM in language understanding and lead to redundant or ineffective tool invocations.

\smallskip
\noindent\textit{Comparison with medical agent.}
CardAIc-Agents also outperformed state-of-the-art medical agents as shown in Table \ref{tab:cp}. Among these, ReConcile showed the largest gap, with accuracies of 0.49 (vs.\ 0.87) on MIMIC-IV, 0.43 (vs.\ 0.96) on PTB-XL, and 0.55 (vs.\ 0.77) on PTB Diagnostic ($p<0.05$). A key limitation of these agents lay in their reliance on the intrinsic capabilities of models; inference remained constrained despite guidance from expert-role prompts. In addition, their static knowledge bases and rigid reasoning pipelines limit adaptation to diverse cases. By contrast, the proposed agent leveraged an updateable CardiacRAG agent, integrated external tools to augment model capabilities, and incorporated an adaptive strategy that enables refinement and optimization across diverse cases.

% Moreover, the sequential nature of their reasoning pipelines could propagate early errors or hallucinations, leading to compounding flaws. By contrast, the proposed agent leveraged an up-to-date CardiacRAG agent, integrated external tools to augment model capabilities, and incorporated reflective mechanisms that continuously revisit prior reasoning steps as well as inputs, thereby reducing error accumulation.

\begin{figure}[t!]
    \centering
    \includegraphics[width=0.75\linewidth]{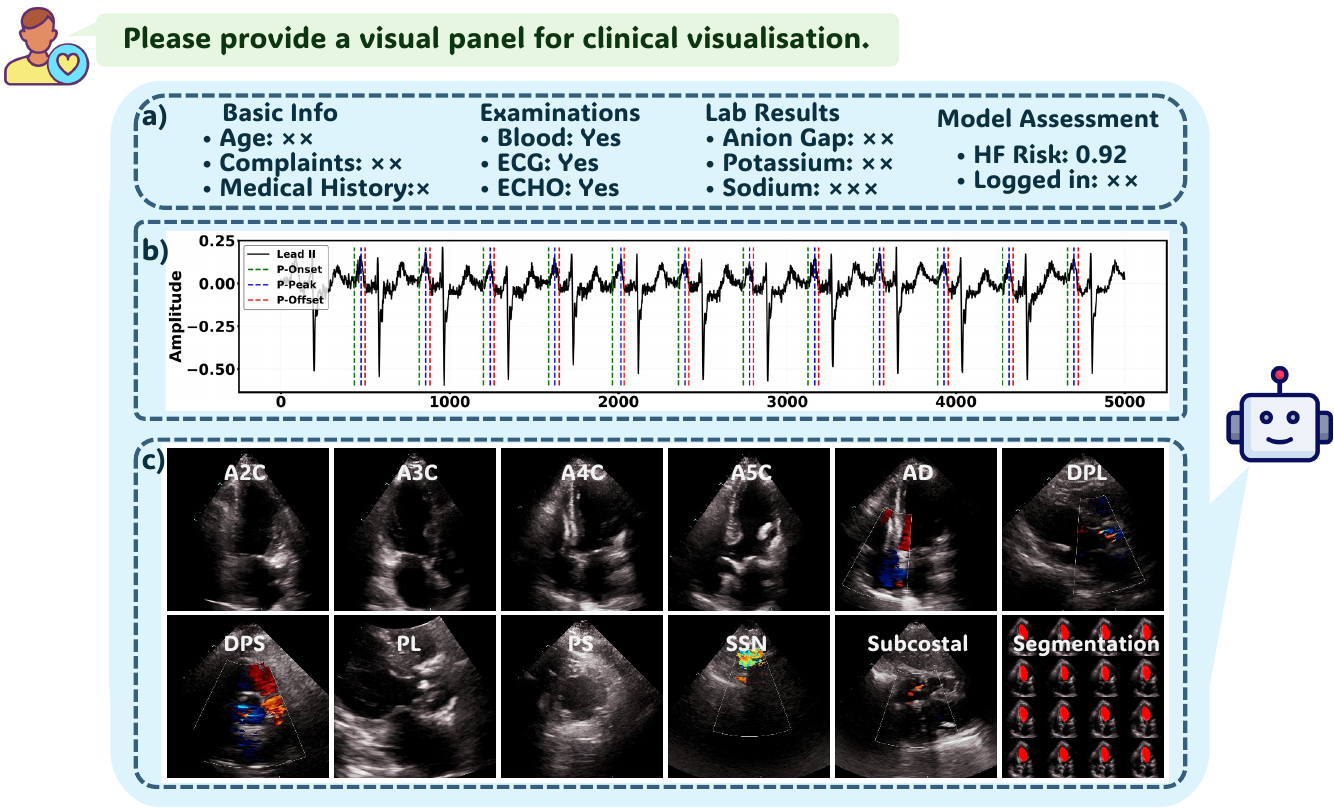}
    \caption{Visual panel generated by CardAIc-Agents: a) patient profile display b) ECG waveform with labeled P and T waves c) echocardiographic view identification with A4C segmentation video frames. Additional details are in the appendices.}
    \label{fig:panel}
    \vspace{-5mm}
\end{figure}

\smallskip
\noindent\textit{Assessment of intermediate visual outputs.}
CardAIc-Agents could provide on-demand support to clinicians for the validation of complex or uncertain cases, a capability enabled by the review panel introduced to facilitate this process (see Figure \ref{fig:panel}). This function was evaluated by two cardiologists. For echocardiography, the agent automatically identified 11 standard views from raw DICOM, achieving 100\% accuracy in key views (eg., A3C, A4C, PLAX, PSAX, SC) and over 80\% accuracy in others (a random sample of 10 cases); Left ventricle segmentation on A4C views had been reported in prior studies with a Dice Coefficient of 0.922 on the EchoNet-Dynamic dataset. Detection of P, QRS, and T waves from 12-lead ECGs was rated suboptimal by experts, mainly due to stringent criteria requiring precise identification of every heartbeat, indicating an area for further improvement. 

\begin{table}[t]
\floatconts
  {tab:ablation}
  {\caption{Ablation Study of Adaptive Workflow, CardiacRAG, and MDT Components}}%
  {%
    \centering
    \vspace{-8pt}
    %\scriptsize
    \setlength{\tabcolsep}{2.5pt}% slightly tighter columns
    \begin{adjustbox}{width=0.75\linewidth}
    \begin{tabular}{c|c|cc|cc|cc}
    \toprule
    \textbf{Level} & \textbf{Workflow} 
     & \multicolumn{2}{c|}{\textbf{CardiacRAG}} 
     & \multicolumn{2}{c|}{\textbf{MDT}} 
     & \textbf{ACC} & \textbf{AUC} \\
    \midrule

    \multirow{3}{*}{Module}
     &  & \multicolumn{2}{c|}{\checkmark} & \multicolumn{2}{c|}{\checkmark} 
       & 0.80 (0.75, 0.85) & 0.87 (0.83, 0.91) \\
     & \checkmark &  && \multicolumn{2}{c|}{\checkmark} 
       & 0.77 (0.72, 0.82)  & 0.81 (0.76, 0.86)  \\
     & \checkmark & \multicolumn{2}{c|}{\checkmark} & & 
       & 0.84 (0.80, 0.88) & 0.88 (0.84, 0.92) \\
    \midrule

    \multirow{4}{*}{\raisebox{-2.5\height}{\begin{tabular}{c}Intra-\\module\end{tabular}}}
     &  & Similarity & Filter & Tool & Memory & ACC & AUC \\
    \midrule
     & \checkmark &  & \checkmark & \checkmark & \checkmark 
       & 0.83 (0.79, 0.87) & 0.87 (0.83, 0.91) \\
     & \checkmark & \checkmark &  & \checkmark & \checkmark 
       & 0.83 (0.78, 0.87) & 0.86 (0.81, 0.90) \\
     & \checkmark & \checkmark & \checkmark &  & \checkmark 
       & 0.75 (0.70, 0.80) & 0.83 (0.78, 0.87) \\
     & \checkmark & \checkmark & \checkmark & \checkmark &  
       & 0.82 (0.77, 0.86) & 0.86 (0.81, 0.90) \\
    \midrule

    Proposed
     & \checkmark & \checkmark & \checkmark & \checkmark & \checkmark
     & \textbf{0.87(0.82,0.90)} & \textbf{0.89(0.85,0.93)} \\
    \bottomrule

    \multicolumn{8}{l}{\normalsize{\checkmark\ = enabled; empty = disabled; Boldface = best performance.}} \\
    \multicolumn{8}{l}{\normalsize{MDT = multidisciplinary discussion team.}} \\
    \end{tabular}
    \end{adjustbox}
  }
  \vspace{-5mm}
\end{table}

\subsection{Ablation Studies}
\noindent\textit{Adaptive workflow.} 
The ablation study was conducted on the MIMIC-IV dataset to evaluate the contribution of the adaptive workflow, where accuracy improved from 0.80 to 0.87 ($p<0.05$, Table \ref{tab:ablation}). This result confirmed the effectiveness of reasoning in an incremental and feedback-aware manner. Specifically, the model performed step-by-step evaluation and summarization, allowing it to re-assess the current state at each stage and adjust the plan accordingly before proceeding. This process emphasized a global plan followed by stepwise adjustments, distinguishing it from a purely ReAct-based approach that prioritizes stepwise changes, as well as from strategies that rely solely on general planning.
% This process not only adapted flexibly to cases but also reduced the risk of error propagation by enabling the system to detect and correct earlier mistakes, rather than compounding them in subsequent reasoning steps.

\smallskip
\noindent\textit{CardiacRAG agent.} The contribution of the CardiacRAG agent is shown in Table \ref{tab:ablation}. The results indicated a clear improvement when a dedicated and independent agent was assigned to generate and refine plans based on domain knowledge, yielding a 10\% performance improvement. This highlighted the effectiveness of the proposed module in precisely retrieving relevant information, as well as the valuable contribution of its curated domain-specific knowledge base. Furthermore, the intra-module ablation of the hybrid retrieval mechanism confirmed that using only vector similarity retrieval or only keyword-based filtering led to decreased performance (more experiments are shown in the appendices). 

\smallskip
\noindent\textit{Multidisciplinary discussion team.} Table \ref{tab:ablation} also reports an increase from 0.84 to 0.87 attributed to the proposed team ($p<0.05$). This gain reflected two drivers: first, the effectiveness of this team to incorporate diverse multimodal information and assign distinct roles to specialized models for collaborative discussion; second, the capability of the agent to dynamically activate the tool upon detecting uncertainty or insufficient evidence in earlier reasoning stages, selectively engaging the team as needed. Such improvement also confirmed the benefits of the proposed adaptation strategy. In addition, intra-module ablation shown that both the tool and its persistent memory contributed to performance, highlighting their importance for stable multi-agent coordination.

\section{Conclusion}
This study introduced CardAIc-Agents, a multimodal framework with adaptive capabilities for cardiac-related tasks. Experiments on three public datasets showed that it outperformed general medical VLMs and state-of-the-art medical agents. In summary, by combining external tools with a cardiac knowledge base, this study presented a hierarchical adaptive framework spanning: complexity assessment; iterative plan refinement as new evidence emerged; dynamic activation of specialized team discussions for complex cases; and provision of visual outputs to support clinician verification. With this adaptive design, CardAIc-Agents delivered scalable multimodal decision support and showed potential for deployment, particularly in resource-limited clinical settings.

\bibliography{main}
\end{document}